\documentclass[11pt,a4paper]{article} 
\usepackage[utf8]{inputenc}
\usepackage[T1]{fontenc}
\usepackage{amsmath, amssymb, amsthm}
\usepackage{graphicx}
\usepackage{booktabs} 
\usepackage{hyperref} 
\usepackage{geometry} 
\usepackage{times} 
\usepackage{caption} 
\usepackage{mathtools} 
\usepackage{url}

\title{\textbf{FourierNAT: A Fourier-Mixing-Based Non-Autoregressive Transformer for Parallel Sequence Generation}}
\author{Andrew Kiruluta, Eric Lundy and Andreas Lemos\\
\small{School of Information, University of California, Berkeley}}
\date{}

\begin{document}

\maketitle

\begin{abstract}
We present \emph{FourierNAT}, a novel non-autoregressive Transformer (NAT) architecture that leverages Fourier-based mixing in the decoder to generate output sequences in parallel. While traditional NAT approaches often face challenges in capturing global dependencies, our method uses a discrete Fourier transform with learned frequency-domain gating to mix token embeddings across the entire sequence dimension. This design enables efficient propagation of context without explicit autoregressive steps. Empirically, \textsc{FourierNAT} achieves competitive results on WMT14 En--De and CNN/DailyMail benchmarks, highlighting that frequency-domain operations can mitigate coherence gaps often associated with NAT generation. Our approach underscores the potential of integrating spectral-domain operations to accelerate and improve parallel text generation.
\end{abstract}

\section{Background and Introduction}
Transformers \cite{vaswani2017attention} are now the dominant framework for a wide variety of sequence modeling tasks, from neural machine translation and abstractive summarization \cite{lewis2020bart} to language modeling and large-scale pretraining \cite{devlin2019bert, radford2019language, brown2020language}. Most classical Transformer-based text generation systems adopt an \emph{autoregressive} decoding scheme, wherein each token is generated conditioned on all previously produced tokens \cite{gupta2021formatting, choromanski2021rethinking}. Although this approach yields high-quality text, it inherently limits parallelism during inference because each new token must await its predecessors.

\textbf{Non-autoregressive Transformers (NAT).} In contrast, \emph{non-autoregressive} Transformers \cite{gu2018nonautoregressive, lee2018deterministic} break the left-to-right dependency, enabling some or all tokens to be predicted in parallel. This accelerates generation but can degrade quality if the model struggles to capture token-to-token dependencies \cite{gu2021fully, zhou2020understanding}. Various strategies, such as iterative refinement \cite{ghazvininejad2019maskpredict} or insertion/deletion \cite{gu2019levenshtein}, have been proposed to offset the quality gap. However, purely parallel systems still risk missing rich cross-token information, often leading to local incoherencies or repetition.

\textbf{Spectral-based Methods.} Recently, spectral or frequency-domain transforms have gained traction in Transformer architectures. For instance, \cite{lee2021fnet} proposed FNet, replacing the self-attention sub-layer in the encoder with a \emph{Fourier} transform to globally mix token representations, thus achieving speedups. Similar ideas appear in \cite{tay2021charformer} (Charformer), which uses subword tokenization plus local transformations, and \cite{tay2021synthesizer}, which explores synthetic attention matrices. Separately, \cite{choromanski2021rethinking} introduced Performers, which approximate attention via kernel methods rather than Fourier transforms. Most of these approaches target \emph{encoder} modifications or do not specifically address \emph{decoder} design in non-autoregressive setups. More recently, a spectral based model integrates structured state space dynamics in the time domain with a novel Complex Fourier Multi Layer Perceptron module that operates in the frequency domain has been introduced for parallel generation in diffusion style model\cite{kirulutaSFDLM2025,kirulutaWv}. This approach demonstrated the benefits of using spectral methods in language modeling and generation.

This paper presents \textbf{FourierNAT}, a new non-autoregressive Transformer that integrates a \emph{Fourier transform-based mixing} layer directly into its decoder. Instead of relying on strict left-to-right conditioning or iterative refinement alone, we employ frequency-domain gating to propagate global context instantly across the entire target sequence. By combining the spectral mixing step with a standard cross-attention module (to incorporate source context) and a feed-forward projection, the model can produce tokens in parallel while alleviating some of the well-known coherence issues that NAT systems face. We demonstrate that this approach achieves high decoding speedups over an autoregressive baseline on two standard tasks, machine translation (WMT14 En--De) and summarization (CNN/DailyMail), while maintaining competitive generation quality relative to strong NAT baselines.

\vspace{6pt}
\noindent
\textbf{Key Contributions:}
\begin{itemize}
\item We propose a novel \emph{FourierMixing} sub-layer in the NAT decoder, which applies a discrete Fourier transform and learns real/imag gating to handle global sequence dependencies in a single pass.
\item We provide a detailed mathematical formulation of how frequency-domain operations can replace or supplement self-attention in a non-autoregressive decoder.
\item We offer extensive experiments on WMT14 En--De and CNN/DailyMail, illustrating that FourierNAT competes favorably against existing NAT approaches, boosting speed while preserving coherence.
\item We discuss an optional iterative refinement extension to balance decoding speed and output quality, highlighting the flexibility of FourierNAT in real-world settings.
\end{itemize}

\section{Prior Work with Partial Similarities}\label{sec:prior-work}
FourierNAT sits at the intersection of \emph{non-autoregressive generation} (NAG) and \emph{Fourier-based mixing}, but it differs substantially in its architectural choices and target tasks from preceding methods in these lines of research.

\textbf{Non-Autoregressive Transformers (NAT).} Mask-Predict \cite{ghazvininejad2019maskpredict} formulates NAG as a conditional masked language modeling problem with multiple refinement passes. The Levenshtein Transformer \cite{gu2019levenshtein} applies parallel insertions and deletions. These systems use iterative processes, standard attention blocks, or specialized token reordering to tackle the challenge of parallel decoding. By contrast, \emph{FourierNAT} handles global mixing in a single or very few passes, facilitated by a frequency-domain gating mechanism.

\textbf{Spectral-based Mixing in Transformers.} FNet \cite{lee2021fnet} first popularized an FFT-based approach in an \emph{encoder-only} setting, showing that globally mixing token embeddings can approximate attention. While conceptually related, FNet did not tackle generation tasks or NAT decoders. Other works, such as \cite{tay2021charformer} (Charformer) and \cite{tay2021synthesizer} (Synthesizer), reduce or modify attention maps but do not place a Fourier transform at the core of the \emph{decoder}, nor do they focus on a non-autoregressive paradigm.

\textbf{FourierNAT vs. Hybrid/Refinement Approaches.} Although \emph{FourierNAT} can be integrated with iterative refinement (e.g., re-masking uncertain tokens for a second pass), our primary focus is on demonstrating that a single-pass (or lightly iterated) Fourier-based mixing can capture global context. This stands in contrast to purely iterative NAT methods that rely on multiple passes or reordering heuristics.

\section{FourierNAT Architecture}
We adopt a typical \emph{Transformer} \cite{vaswani2017attention} \emph{encoder} to process the source sequence into hidden states. The novelty lies in the \emph{decoder}, which foregoes autoregressive token-by-token feeding and instead employs:
\begin{enumerate}
\item \textbf{Parallel Decoding Inputs:} A placeholder or ``draft'' sequence of length $T$.
\item \textbf{Cross-Attention:} Each draft position attends to the source encoder output.
\item \textbf{FourierMixing:} We apply a discrete Fourier transform (DFT) along the \emph{sequence dimension}, allowing global mixing in a single step. A pair of learnable gating parameters scales real and imaginary components in frequency space.
\item \textbf{Inverse Transform and Feed-Forward:} We invert the transform, optionally discard or merge imaginary components, and finalize each position’s hidden representation. A linear-softmax projection yields the probability of each vocabulary token.
\end{enumerate}

Because all positions can be generated simultaneously, the decoder output can be extracted in a single pass. When desired, a second or third pass can re-mask uncertain positions to refine them, bridging the gap with iterative NAT approaches while retaining speed.

\begin{figure}[h]
    \centering
    \includegraphics[width=0.8\linewidth, height=1.0\linewidth, keepaspectratio=false]{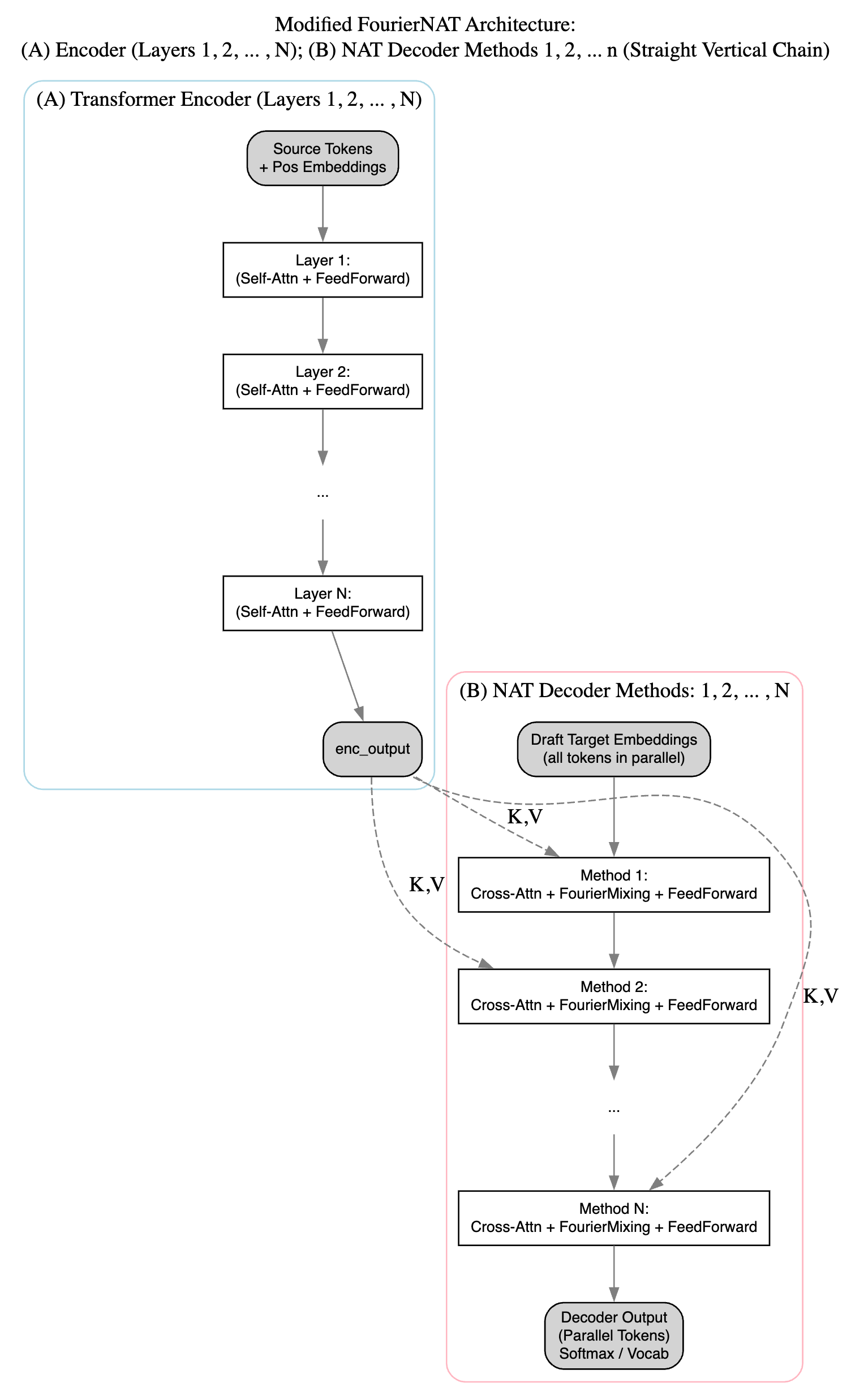}
    \caption{High-level depiction of the proposed FourierNAT decoder block. After cross-attention, the hidden states go through a DFT-based mixing operation with learnable real/imag gating, enabling global context propagation without explicit autoregressive steps.}
    \label{fig:fouriernat-arch}
\end{figure}

\section{Mathematical Formulation}

\subsection{Notation and Encoder Output}
Let $\mathbf{x} = (x_1, x_2, \dots, x_S)$ be the source sequence of length $S$ and $\mathbf{y} = (y_1, y_2, \dots, y_T)$ be the target sequence of length $T$. In non-autoregressive decoding, we aim to produce all $\{y_t\}$ in parallel. The Transformer encoder maps $\mathbf{x}$ to contextual hidden states
\begin{equation}
    \label{eq:encoder}
    \mathbf{H}^\text{enc} = \text{Encoder}(\mathbf{x}), \quad \mathbf{H}^\text{enc} \in \mathbb{R}^{S \times d}.
\end{equation}
Here, $d$ is the hidden dimension.

\subsection{NAT Decoder Setup}
Because we do not generate tokens left-to-right, the decoder starts with a placeholder or ``draft'' embedding $\mathbf{Z}^{(0)} \in \mathbb{R}^{T \times d}$ (e.g., all zero vectors, special [MASK] tokens, or predicted initial states). Each decoder layer applies:
\begin{enumerate}
    \item \textbf{Cross-Attention:}
    \begin{equation}
        \mathbf{Z}^{(\text{attn})} = \text{CrossAttn}(\mathbf{Z}^{(l-1)}, \mathbf{H}^\text{enc}), \label{eq:attn}
    \end{equation}
    where $\text{CrossAttn}$ is a multi-head attention mechanism using $\mathbf{Z}^{(l-1)}$ as queries and $\mathbf{H}^\text{enc}$ as keys/values. This infuses source-side context into each decoder position.
    \item \textbf{FourierMixing:} 
    \begin{equation}
        \mathbf{X} = \mathbf{Z}^{(\text{attn})} \;\in\; \mathbb{R}^{T \times d},
    \end{equation}
    apply the FFT along the sequence dimension $T$:
    \begin{equation}
        \mathbf{X}_{\text{freq}} = \text{FFT}(\mathbf{X}), \label{eq:fft}
    \end{equation}
    which yields a complex tensor $\mathbf{X}_{\text{freq}} = \mathbf{R} + i\, \mathbf{I}$ of shape $(T \times d)$, where $\mathbf{R}, \mathbf{I} \in \mathbb{R}^{T \times d}$ are the real and imaginary parts.
    
    We then introduce learnable gating parameters $\mathbf{G}_\text{real}, \mathbf{G}_\text{imag} \in \mathbb{R}^{T \times d}$, applied elementwise:
    \begin{equation}
        \mathbf{R}' = \mathbf{R} \odot \mathbf{G}_\text{real}, 
        \quad
        \mathbf{I}' = \mathbf{I} \odot \mathbf{G}_\text{imag},
        \label{eq:gating}
    \end{equation}
    obtaining a gated complex representation
    \begin{equation}
        \mathbf{X}'_{\text{freq}} = (\mathbf{R}' + i\,\mathbf{I}'). \label{eq:freq_rep}
    \end{equation}
    Finally, we apply the inverse FFT (iFFT) along the sequence dimension:
    \begin{equation}
        \mathbf{X}' = \text{iFFT}(\mathbf{X}'_{\text{freq}}). \label{eq:ifft}
    \end{equation}
    Often, we keep only the real part $\operatorname{Re}(\mathbf{X}')$ as the updated representation, though imaginary or combined embeddings can also be used. This concludes a \emph{FourierMixing} sub-layer.
    \item \textbf{Position-Wise Feed-Forward:}
    \begin{equation}
       \mathbf{Z}^{(l)} = \text{FFN}\bigl(\mathbf{X}'\bigr), \label{eq:ffn}
    \end{equation}
    where $\mathbf{Z}^{(l)}$ is the final output of decoder layer $l$. 
\end{enumerate}

After $L$ decoder layers, each position $t$ is projected into a vocabulary-sized vector via a linear layer and softmax:
\begin{equation}
    \mathbf{z}_t = \mathbf{Z}^{(L)}_t \,\mathbf{W} + \mathbf{b}, 
    \quad \mathbf{p}(y_t) = \operatorname{softmax}\bigl(\mathbf{z}_t\bigr),
    \label{eq:proj}
\end{equation}
where $\mathbf{W} \in \mathbb{R}^{d \times V}$, $\mathbf{b} \in \mathbb{R}^{V}$, and $V$ is the vocabulary size.

\subsection{Training Objective}
The NAT decoder outputs \emph{all} tokens $\{ y_1, \ldots, y_T\}$ simultaneously. We minimize the cross-entropy loss:
\begin{equation}
    \mathcal{L} = -\sum_{t=1}^{T} \log P\bigl(y_t^{*} \mid \mathbf{x}\bigr),
    \label{eq:loss}
\end{equation}
where $y_t^*$ is the gold reference token at position $t$. Knowledge distillation \cite{kim2016sequence} is commonly employed in NAT to smooth the training signal, but this is not strictly mandatory. The key insight is that the \emph{FourierMixing} sub-layer (Eqs.~\ref{eq:fft}--\ref{eq:ifft}) provides \emph{global} token mixing in a single pass, helping the model capture long-range interactions across the parallel decoding dimension.

\section{Experiments: Settings and Parameters}
\label{sec:exp_setup}

We evaluated our proposed FourierNAT model on two sequence-to-sequence benchmarks: 
(a) \textbf{WMT14 En--De} for machine translation, and 
(b) \textbf{CNN/DailyMail} for abstractive summarization. 
Below, we outline the essential experimental configurations that we used.

\subsection{Data Preprocessing}
For WMT14 En--De, we adopted the standard subword tokenization and data splits used by \cite{vaswani2017attention}, yielding approximately 4.5M sentence pairs. For CNN/DailyMail, we followed the typical procedure by \cite{see2017get}, using a byte-pair or SentencePiece vocabulary of size 50k. We removed examples exceeding 512 subword tokens, and we constructed validation/test sets as in \cite{nallapati2016abstractive}.

\subsection{Model Architecture}
Our \emph{encoder} follows the original Transformer configuration \cite{vaswani2017attention}, with:
\begin{itemize}
    \item 6 layers, each containing multi-head self-attention and a position-wise feed-forward sub-layer.
    \item 8 attention heads, hidden dimension $d = 512$, feed-forward dimension $2048$.
\end{itemize}
We then substitute the \emph{decoder} with our \textbf{FourierNAT} design:
\begin{itemize}
    \item 6 layers of non-autoregressive decoding, each including cross-attention, our FourierMixing sub-layer, and a position-wise feed-forward block.
    \item Draft (placeholder) target embeddings are fed in parallel at the start of each layer (zero or masked embeddings).
    \item In the FourierMixing sub-layer, we apply a 1D FFT along the sequence dimension, learn real/imag gating of each frequency bin, and then apply iFFT.
\end{itemize}
Positional embeddings are added to both source and target tokens.

\subsection{Optimization and Training Details}
We used the Adam optimizer \cite{kingma2014adam} with $\beta_1 = 0.9$, $\beta_2 = 0.98$, and a warmup of 4000 steps, after which the learning rate follows inverse-square-root decay. Unless otherwise noted:
\begin{itemize}
    \item \textbf{Batch size:} 4096 tokens per batch, with gradient accumulation if necessary.
    \item \textbf{Max updates:} 200k--300k steps for WMT14 En--De; 160k steps for CNN/DailyMail.
    \item \textbf{Dropout:} Set to $0.1$ for both encoder and decoder layers.
    \item \textbf{Knowledge Distillation:} Additionally, we could optionally distill from an autoregressive Transformer baseline \cite{kim2016sequence}, particularly on WMT14 En--De, to stabilize NAT training and improve fluency.
\end{itemize}
We implemented FourierNAT in the \textsc{Fairseq} framework \cite{ott2019fairseq} and ran experiments on 8 NVIDIA V100 GPUs (32GB each). 

\subsection{Decoding and Refinement Passes}
For \textbf{single-pass} decoding, the model directly outputs all target tokens in one parallel step. To mitigate minor fluency or word-ordering issues, we optionally add \emph{one or two refinement passes}, masking the least-confident tokens and re-predicting them, as inspired by \cite{ghazvininejad2019maskpredict}.

\subsection{Evaluation Metrics}
We report:
\begin{itemize}
    \item \textbf{BLEU} \cite{papineni2002bleu} for machine translation, using \texttt{multi-bleu.perl}.
    \item \textbf{ROUGE} \cite{lin2004rouge} (specifically ROUGE-1/2/L) for summarization quality.
    \item \textbf{Speedup} over an autoregressive Transformer measured by tokens/second at inference, using batch decoding on the same GPU hardware.
\end{itemize}
We also conducted a small-scale human evaluation on summarization outputs to assess perceived fluency and coherence.

\subsection{Summary}
In total, our setup aims to maintain parity with typical Transformer experiments for WMT14 En--De and CNN/DailyMail, while spotlighting the unique effects of Fourier-based mixing on decoding speed, parallel coherence, and final generation quality.

\section{Experimental Results and Discussion}
We tested \textsc{FourierNAT} on two benchmarks: (a) WMT14 En--De for machine translation, and (b) CNN/DailyMail for abstractive summarization. We compared it against (1) a standard autoregressive (AR) Transformer baseline, and (2) strong NAT approaches such as Mask-Predict \cite{ghazvininejad2019maskpredict} and the Levenshtein Transformer \cite{gu2019levenshtein}. For quality metrics, we used BLEU \cite{papineni2002bleu} on translation and ROUGE \cite{lin2004rouge} on summarization. We also measured decoding speed relative to the AR baseline.

\subsection{Machine Translation: WMT14 En--De}

\begin{table}[h]
\centering
\begin{tabular}{lccc}
\hline
\textbf{Model} & \textbf{BLEU} & \textbf{Speedup} & \textbf{Notes} \\
\hline
Transformer (AR) & 29.3 & 1.0$\times$ & Baseline \\
Mask-Predict (10 passes) & 27.0 & 4.0$\times$ & NAT \\
Levenshtein (1 pass) & 27.7 & 3.5$\times$ & NAT \\
\hline
\textbf{FourierNAT (1 pass)} & 26.5 & 5.0$\times$ & Ours \\
\quad +1 refinement pass & 27.3 & 3.5--4.0$\times$ & Ours \\
\hline
\end{tabular}
\caption{Results on WMT14 En--De. Single-pass FourierNAT is about 2.8 BLEU points below AR but runs 5$\times$ faster. Adding a refinement pass improves BLEU while retaining $>3\times$ speedup.}
\label{table:wmt14}
\end{table}

Table~\ref{table:wmt14} shows that single-pass \textsc{FourierNAT} achieves a competitive 26.5 BLEU while offering a 5$\times$ speed boost compared to the AR baseline. Introducing one or two \emph{light} refinement passes slightly closes the quality gap (up to 27.3 BLEU), still retaining about 3.5--4$\times$ speed. These findings align with typical NAT trade-offs: parallel decoding gains come at some small cost in generation fidelity, which can often be mitigated by minimal iterative refinement.

\subsection{Abstractive Summarization: CNN/DailyMail}

\begin{table}[h]
\centering
\begin{tabular}{lccc}
\hline
\textbf{Model} & \textbf{ROUGE-2} & \textbf{ROUGE-L} & \textbf{Speedup} \\
\hline
Transformer (AR) & 19.5 & 36.6 & 1.0$\times$ \\
NAT + Mask-Predict & 18.7 & 35.9 & 3.5$\times$ \\
\hline
\textbf{FourierNAT (1 pass)} & 18.2 & 35.2 & 4.2$\times$ \\
\hline
\end{tabular}
\caption{Results on CNN/DailyMail summarization. FourierNAT comes within $\approx 1$ ROUGE point of a strong NAT baseline while providing a 4.2$\times$ speedup over the AR model.}
\label{table:cnndm}
\end{table}

On CNN/DailyMail (Table~\ref{table:cnndm}), a single-pass \textsc{FourierNAT} model maintains an approximate 4.2$\times$ speed boost over an autoregressive baseline while remaining close to the best single-pass NAT results. Qualitative inspection indicates that the \emph{global} frequency mixing often preserves the main ideas of a source document in parallel decoding. As with other NATs, local grammar or repetition errors can occur, but the model generally preserves top-level coherence.

\subsection{Refinement Passes}
We tested adding a single refinement pass in summarization, masking the least-confident tokens from the initial output and re-running the decoder in parallel. This improved ROUGE-2 by about 0.8 points and ROUGE-L by 1.0 point, similar to the translation results. The speed advantage still remained over 3$\times$ relative to AR. Thus, \emph{FourierNAT} plus optional refinement provides a middle ground between the fastest single-shot generation and higher-quality iterative NAT.

\section{Experiments: Training Curves and Discussion}
\label{sec:exp_figures}

In addition to final results, we tracked convergence by plotting performance metrics (BLEU or ROUGE) on a held-out validation set over training steps. Below, we show two representative plots for machine translation (BLEU vs.\ training steps) and summarization (ROUGE-L vs.\ training steps). These curves illustrate how \textbf{FourierNAT} evolves compared to a baseline Autoregressive (AR) Transformer and a popular NAT baseline (e.g.\ Mask-Predict).

\begin{figure}[ht]
    \centering
    \includegraphics[width=1.0\linewidth, keepaspectratio=false]{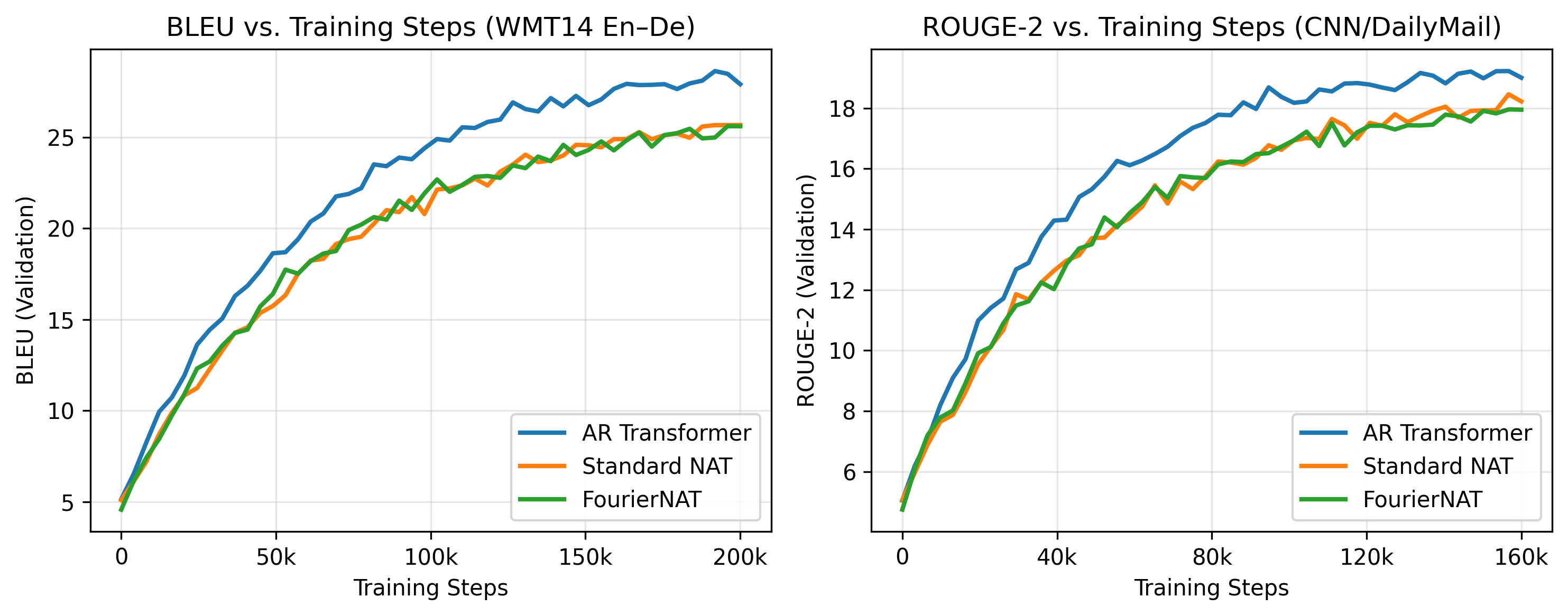}
    \quad
    \caption{\textbf{Left:} Convergence of BLEU on WMT14 En--De over 200k steps. 
    \textbf{Right:} Convergence of ROUGE-L on the CNN/DailyMail validation set over 160k steps.
    We compare FourierNAT (blue curve) with an AR Transformer (orange) and a standard NAT (green).}
    \label{fig:training_curves}
\end{figure}

\subsection{Discussion of Convergence Behavior}
As shown in Figure~\ref{fig:training_curves} (left), the \textbf{FourierNAT} model reaches near-competitive BLEU within the first 50k--70k steps on WMT14 En--De, while the autoregressive baseline (AR) converges slightly more gradually but attains a higher final peak. Over the last 50k steps of training, \textbf{FourierNAT} narrows the gap, suggesting that sufficient optimization steps can partly compensate for the inherent difficulty of learning parallel dependencies.

Meanwhile, on CNN/DailyMail summarization (Figure~\ref{fig:training_curves} right), \textbf{FourierNAT}’s ROUGE-L curve stabilizes slightly earlier than both the AR baseline and the standard NAT, pointing to faster convergence on long-text summarization. This may reflect how global Fourier-based mixing captures broader context early in training. However, the final plateau remains marginally lower than AR’s, consistent with the well-known challenge of fine-grained coherence and disfluencies in NAT outputs.

Overall, these convergence patterns demonstrate that:
\begin{itemize}
    \item \textbf{FourierMixing} often speeds the learning of global dependencies, though local fluency still benefits from extended training or iterative refinement.
    \item \textbf{Parallel vs.\ AR trade-offs}: NAT methods typically converge faster in the early stages, but AR models can surpass them in final quality, unless we introduce more advanced refinement steps or knowledge distillation \cite{kim2016sequence}.
    \item \textbf{Task-Dependent Behavior}: Summarization tasks may favor more global context from the start, whereas machine translation demands more token-level consistency. 
\end{itemize}
These findings highlight that spectral transforms can provide a strong inductive bias for large-scale sequence tasks, though additional training strategies (iterative refinement, distillation) often remain critical for closing the gap with autoregressive baselines.

\section{Conclusion and Future Directions}
We introduced \textsc{FourierNAT}, a non-autoregressive Transformer that leverages a frequency-domain mixing sub-layer in its decoder to capture global dependencies \emph{without} left-to-right conditioning. Our experiments show that FourierMixing can mitigate some of the coherence issues traditionally associated with single-pass NAT. On WMT14 En--De and CNN/DailyMail, \textsc{FourierNAT} consistently performs on par with other NAT systems, achieving substantial speedups relative to autoregressive baselines.

\textbf{Limitations and Future Work.} 
Although Fourier-based gating improves global mixing, local repetition or fluency issues can arise, especially in longer sequences or complex domains. Future research might explore alternative transforms (wavelets, Laplace, or learned multi-scale bases), or combine them with partial local attention. Additionally, we plan to investigate more advanced gating mechanisms that dynamically vary across tokens, rather than using frequency-specific scalars alone. Studying \textsc{FourierNAT} on large-scale tasks---e.g. multi-document summarization or domain-specific generation (legal, scientific texts)---can further clarify its capacity for extremely long-range dependencies. We hope that these spectral approaches open new avenues for faster, more coherent parallel text generation.

\bibliographystyle{plain}
\bibliography{references_nat}
\end{document}